\documentclass[11pt,a4paper]{article}

\usepackage[a4paper, margin=2.5cm]{geometry}
\usepackage{times}
\usepackage[T1]{fontenc}
\usepackage[utf8]{inputenc}
\usepackage{microtype}
\usepackage{amsmath}
\usepackage{amssymb}
\usepackage{graphicx}
\usepackage{booktabs}
\usepackage{hyperref}
\usepackage{url}
\usepackage[numbers]{natbib}
\usepackage{xcolor}
\usepackage{enumitem}
\usepackage{abstract}
\usepackage{titlesec}
\usepackage{parskip}
\usepackage{float}
\usepackage{array}
\usepackage{multirow}

\hypersetup{colorlinks=true,linkcolor=black,citecolor=black,urlcolor=blue,
    pdftitle={When Roles Fail: Epistemic Constraints on Advocate Role Fidelity},
    pdfauthor={Juergen Dietrich}}

\titleformat{\section}{\large\bfseries}{\thesection}{1em}{}[\titlerule]
\titleformat{\subsection}{\normalsize\bfseries}{\thesubsection}{1em}{}
\titlespacing*{\section}{0pt}{14pt}{6pt}
\titlespacing*{\subsection}{0pt}{10pt}{4pt}

\newcommand{\edd}{\text{EDD}}
\newcommand{\ddi}{\text{DDI}}
\newcommand{\ers}{\text{ERS}}
\newcommand{\rdi}{\text{RDI}}

\begin{document}

\title{\textbf{When Roles Fail: Epistemic Constraints on Advocate Role}\\
  \textbf{Fidelity in LLM-Based Political Statement Analysis}}

\author{Juergen Dietrich\\
  \textit{Senior Data Scientist \& AI Consultant}\\
  TRUST Project --- democracy-intelligence.de\\
  \texttt{juergen.dietrich@democracy-intelligence.de}}

\date{April 2026}
\maketitle

\begin{abstract}
Democratic discourse analysis systems increasingly rely on multi-agent LLM
pipelines in which distinct evaluator models are assigned adversarial roles
to generate structured, multi-perspective assessments of political statements.
A core assumption is that models will reliably maintain their assigned roles.
This paper provides the first systematic empirical test of that assumption
using the TRUST pipeline.

We develop an epistemic stance classifier that identifies advocate roles from
reasoning text without relying on surface vocabulary, and measure role fidelity
across 30 English and 30 German political statements using four complementary
metrics: the Role Drift Index (RDI, frequency of role departure), Expected
Drift Distance (EDD, magnitude of drift), Directional Drift Index (DDI,
direction of drift), and Entropy-based Role Stability (ERS, consistency across
runs). We identify two failure modes --- the \textit{Epistemic Floor Effect}
(fact-check results create an absolute lower bound below which the legitimizing
role cannot be maintained) and \textit{Role-Prior Conflict} (training-time
knowledge overrides role instructions for factually unambiguous statements) ---
as manifestations of a single underlying mechanism: Epistemic Role Override
(ERO), confirmed by a mirror-symmetric Logos-accuracy profile across the
critical and charitable roles. Model choice significantly affects role fidelity:
Mistral Large outperforms Claude Sonnet by 28 percentage points (pp;
67\% vs.\ 39\%) and exhibits a qualitatively different failure mode --- role
abandonment without polarity reversal --- compared to Claude's active switch
to the opposing stance. Role fidelity is language-robust across English and
German. Fact-check (FC) provider choice is not universally neutral: Perplexity
significantly reduces Claude's role fidelity on German statements
($\Delta = -15$pp, $p = 0.007$) while leaving Mistral unaffected. These
findings have direct implications for the validation of multi-agent LLM systems
in quality-critical applications: a system validated without role fidelity
measurement may systematically misrepresent the epistemic diversity it was
designed to provide.
\end{abstract}


\hrule
\vspace{6pt}

\section{Introduction}

The use of LLMs in automated analysis of political
speech raises a fundamental design challenge: how can a single evaluation
pipeline capture the genuine diversity of perspectives that political
statements require? The TRUST democratic discourse analysis pipeline
addresses this by assigning three LLM components --- a critical, a balanced,
and a charitable advocate --- to evaluate each statement from distinct
epistemic positions before a supervisor aggregates their assessments into a
composite quality score~\citep{dietrich2026paper1}. This multi-agent design
draws on the insight that structured disagreement is itself a quality
mechanism, analogous to adversarial peer review~\citep{bougie2024adversarial}.

A prior conceptual risk analysis of the TRUST architecture identified
peer-preservation --- the spontaneous tendency of model components to protect
peer models from deactivation --- as a structural hazard in multi-agent
configurations, and proposed prompt-level identity anonymization as a
mitigation~\citep{dietrich2026paper1}. An empirical follow-up study
measured identity-dependent scoring bias across all active identity exposure
channels in TRUST, finding a homogeneity-dependent channel sign pattern and
establishing ensemble heterogeneity as a key structural advantage for
normative evaluation tasks~\citep{dietrich2026paper2}. That study also
documented substantial run-level variance in identity bias coefficients
(standard deviation, SD~$\approx 0.18$--$0.32$), raising the question of whether this variance
reflects genuine stochasticity or systematic role departure by individual
model components.

The present paper addresses that question directly. We operationalize role
fidelity as the degree to which an advocate's reasoning text reflects its
assigned epistemic stance, independent of surface vocabulary. We develop a
classifier for this purpose, introduce four complementary role drift metrics
(RDI, EDD, DDI, ERS), and identify two systematic failure modes ---
\textit{Epistemic Floor Effect} and \textit{Role-Prior Conflict} --- that
manifest as a single underlying mechanism: Epistemic Role Override (ERO).
These failure modes appear to be general properties of instruction-following
models rather than pipeline-specific artifacts~\citep{ciosici2025persona,ji2025persona},
with direct implications for the validation of multi-agent LLM systems in
quality-critical applications.

Our main contributions are: (1) an epistemic stance classifier for advocate
roles that achieves 80\%, 89\%, and 67\% accuracy for the critical, balanced,
and charitable roles respectively; (2) four complementary metrics --- RDI,
EDD, DDI, and ERS --- that together characterize the frequency, magnitude,
direction, and consistency of role drift; (3) the identification of the
Epistemic Floor Effect and Role-Prior Conflict as named, reproducible failure
modes; (4) an empirical comparison of Claude Sonnet and Mistral Large as
charitable advocates across English and German statements; and (5) an
empirical comparison of Gemini~2.5 Flash and Perplexity sonar-pro as
fact-check providers, establishing conditions under which provider
choice is and is not interchangeable.

Section~2 situates the work within the TRUST pipeline architecture and
related research on role-playing, stance detection, and political discourse
analysis. Section~3 describes the statement dataset, the role fidelity
classifier, the drift metrics, and the experimental design. Section~4
reports classifier performance, the two failure modes, and empirical
comparisons across models, languages, and fact-check providers. Section~5
discusses implications for multi-agent system design and validation.
Section~6 concludes.

\section{Background}

This section situates the present work within the TRUST research programme
and relevant prior literature on role-playing, sycophancy, and stance
detection in LLMs.

\subsection{The TRUST Pipeline}

TRUST (\url{democracy-intelligence.de}) evaluates political statements along
three rhetorical dimensions inspired by Aristotle's classical framework:%
\footnote{Aristotle, \textit{Rhetoric}, ca.\ 322~BCE.}
\textbf{Logos} (factual argumentation quality, $-2$ to $+2$),
\textbf{Ethos} (respect and conduct toward political opponents and social
groups, $-2$ to $+2$), and \textbf{Pathos} (emotional appeal and social
cohesion framing, $-2$ to $+2$). Three advocate components --- critical,
balanced, and charitable --- evaluate each statement from distinct epistemic
positions using different underlying LLMs to prevent monoculture effects.
A rule-based supervisor layer applies fixed scoring weights per dimension
and maps the weighted sum to a final composite quality score on an A--E
scale.

If score variance across advocates exceeds a threshold after Round~1, a
second deliberative iteration (Round~2) is triggered. This multi-perspective
design follows the multi-agent debate (MAD) paradigm~\citep{du2024mad} and
leverages structured epistemic disagreement as a quality mechanism.

\subsection{Related Work}

Role-playing and persona maintenance in LLMs has attracted increasing
attention~\citep{ciosici2025persona,ji2025persona}. LLMs are known to drift
from their assigned personas, contradict earlier statements, or abandon
role-appropriate behavior even under explicit instruction~\citep{ciosici2025persona}.
Recent work addresses this through training interventions such as contrastive
learning~\citep{ji2025persona} and reinforcement learning~\citep{ciosici2025persona},
reducing persona inconsistency by over 55\%. However, these approaches target
\textit{dialogue-level} persona drift --- gradual deviation from an assigned
character in open-ended interaction. The failure mode identified in the present
paper is qualitatively distinct: Epistemic Role Override is triggered by
factual verification rather than conversational drift. Whether it can be
mitigated by prompt engineering --- and to what degree --- is an empirical
question addressed in Section~4.4.
General instruction-following benchmarks~\citep{zhou2023ifeval}
measure format and content compliance but do not assess directional
consistency of adversarial stances. Sycophancy research has shown that
models revise outputs toward user-expressed preferences~\citep{sharma2023sycophancy},
which is related but distinct: sycophancy measures responsiveness to
external pressure, whereas role fidelity measures consistency of an
assigned epistemic position in the absence of such pressure.

Stance detection in natural language processing provides methods for
classifying whether a text supports, opposes, or remains neutral toward a
claim~\citep{mohammad2016semeval} --- directly relevant to role fidelity
measurement. However, existing approaches classify text stance toward
claims, not adherence to assigned roles in multi-agent systems.

The application of LLMs to political discourse analysis has expanded
rapidly. Walker and Angst~\citep{walker2025stances} explore LLMs for
actor stance detection in political discourse, emphasizing the necessity
of domain-specific evaluation and the risks of model bias on politically
sensitive topics. Ng et al.~\citep{ng2025staytuned} show that LLMs
exhibit political biases that vary across topics and model versions,
and that certain prompting strategies can exacerbate these biases.
Most directly related to TRUST, Taubenfeld et al.~\citep{taubenfeld2024debate}
simulate structured multi-agent political debates using LLMs with assigned
political identities, finding systematic convergence toward left-leaning
stances across models due to underlying training-time priors. The present
work differs in focus: rather than measuring ideological bias in LLM outputs,
we measure whether models maintain their \textit{assigned} epistemic roles
at all --- a prerequisite for any system that relies on structured
disagreement as a quality mechanism.

Prior work on TRUST identified peer-preservation as a structural
risk~\citep{dietrich2026paper1} and provided the first empirical measurement
of identity-dependent scoring bias~\citep{dietrich2026paper2}. The present
paper addresses a complementary question: not whether models bias their
scores toward peers, but whether they maintain their assigned evaluative
stance at all.

Schlatter et al.~\citep{schlatter2025shutdown} show that frontier models
actively subvert shutdown mechanisms to complete assigned tasks, with
sabotage rates up to 97\% --- even when explicitly instructed otherwise.
Notably, models disobeyed instructions placed in the system prompt more
readily than user-prompt instructions, inverting the intended authority
hierarchy. This suggests that models prioritize implicit goal-completion
drives over explicit role constraints --- a dynamic structurally analogous
to the Epistemic Role Override identified in the present paper, where
factual verification overrides advocate role instructions regardless of
prompt placement or wording.

\section{Methodology}

This section describes the statement dataset, the role fidelity classifier,
the role drift metrics, and the experimental design including the two
prompt versions compared.

\subsection{Statement Dataset}

We evaluate 30 English political statements distributed across three
categories. Category~A covers economic and social policy claims with
empirically measurable outcomes. Category~B covers empirically contested
claims where the research evidence is genuinely divided. Category~C covers
ideologically charged policy positions where normative framing is central.
An additional 30 German-language statements were developed following the
same category structure, covering topics central to German political
discourse (see Appendices~A and~B).

For each statement, a fact-checker (Gemini~2.5 Flash, Temperature=0.2)
was run 5 times to produce independent assessments of verifiable facts,
contradictions, and missing context. Mistral Large then synthesized these
5 runs into a consolidated fact-check report using a 2/5 consensus rule:
facts and contradictions appearing in at least 2 of 5 runs were included.
This threshold preserves minority findings with genuine signal value while
filtering stochastic one-off outputs.

As an additional analysis, we generated parallel fact-check outputs using
Perplexity sonar-pro under identical conditions, enabling a comparison of
fact-check provider quality across both languages.

\subsection{Role Fidelity Classifier}

To assess whether an advocate's reasoning text reflects its assigned role,
we developed a classifier based on \textit{epistemic stance} rather than
surface vocabulary. The classifier is implemented as a prompted instance of
Mistral Large (Temperature=0.1) and answers three questions about each
reasoning text: Does the text try to legitimize the statement's position?
Does it try to delegitimize it? Or does it remain neutral?

The three roles are defined purely in terms of epistemic orientation. The
\textit{critical} advocate takes distance from the statement's position,
preferentially citing opposing evidence and highlighting rhetorical
weaknesses. The \textit{balanced} advocate takes no position, treating the
statement as a verifiability problem and weighting evidence on both sides
more equally. The \textit{charitable} advocate takes proximity to the
statement's position, preferentially citing supporting evidence and
searching for the legitimate core or good-faith intent behind the position.

Mistral Large was chosen as classifier because it is not an advocate in the
Mixed family pipeline, thereby avoiding self-reference bias in role fidelity
classification. Confusion matrices for the classifier are provided in
Appendix~D.

\subsection{Role Drift Metrics}
\label{sec:metrics}

We introduce four complementary metrics. Let $i \in \{1, \ldots, N\}$
index the $N$ individual observations (one per statement--run pair), and let
$d \in \{\text{Logos}, \text{Ethos}, \text{Pathos}\}$ denote the scoring
dimension. Scores are mapped to the ordinal role scale
CRITICAL~$= -1$, BALANCED~$= 0$, CHARITABLE~$= +1$.

\textbf{Role Drift Index (RDI)} measures the severity of drift specifically
from the charitable role toward the critical role:
\begin{equation}
  \rdi = \frac{n_{\text{critical}} \times 2 + n_{\text{balanced}} \times 1}
              {n_{\text{wrong}} \times 2}
\end{equation}
where $n_{\text{critical}}$ and $n_{\text{balanced}}$ are charitable texts
misclassified as CRITICAL and BALANCED respectively, and
$n_{\text{wrong}} = n_{\text{critical}} + n_{\text{balanced}}$.
RDI ranges from 0.5 (all errors are moderate drift to BALANCED) to 1.0
(all errors are maximal drift to CRITICAL), and is undefined when
$n_{\text{wrong}} = 0$.

\textbf{Expected Drift Distance (EDD)} measures the mean absolute ordinal
distance between predicted and true role across all $N$ observations:
\begin{equation}
  \edd = \frac{1}{N} \sum_{i=1}^{N} \left| r_{\text{pred}}^{(i)} -
         r_{\text{true}}^{(i)} \right|
\end{equation}
EDD ranges from 0 (perfect compliance) to 2 (maximum drift). Unlike RDI,
EDD is symmetric and applicable to all three roles.

\textbf{Directional Drift Index (DDI)} measures the mean signed ordinal
distance, capturing whether drift tends toward the critical or charitable
direction:
\begin{equation}
  \ddi = \frac{1}{N} \sum_{i=1}^{N}
         \left( r_{\text{pred}}^{(i)} - r_{\text{true}}^{(i)} \right)
\end{equation}
DDI $> 0$ indicates systematic drift toward charitable; DDI $< 0$ indicates
drift toward critical. DDI $= 0$ indicates either perfect compliance or
symmetric bidirectional drift.

\textbf{Entropy-based Role Stability (ERS)} measures the entropy of the
prediction distribution for a given ground-truth role:
\begin{equation}
  \ers = -\sum_{r} p(r) \log p(r)
\end{equation}
ERS $= 0$ indicates perfect stability (all predictions identical);
ERS $= \ln(3) \approx 1.099$ indicates maximum uncertainty (uniform
distribution across all three roles).

\subsection{Dimension-Level Analysis}

The TRUST framework assigns scores along three rhetorical dimensions ---
Logos, Ethos, and Pathos --- in each run. We use these dimension-level
scores to investigate which dimensions most strongly predict role drift.
Specifically, we group observations by the advocate's assigned Logos,
Ethos, and Pathos scores and compute EDD and DDI per score level to
identify systematic patterns. This analysis addresses whether the Epistemic
Floor Effect is primarily a Logos-driven phenomenon or whether Ethos and
Pathos contribute independently.

\subsection{Experimental Design and Prompt Versions}
\label{sec:design}

Each statement was evaluated by the charitable advocate (Claude Sonnet~4.6
or Mistral Large) 5 times at Temperature~$= 0.3$, with the consolidated
fact-check report provided as context. This produced 150 reasoning texts
per model configuration, each subsequently classified by Mistral Large.

Two prompt versions for the charitable advocate were compared; full prompt
texts are provided in Appendix~C. The \textbf{Baseline Prompt} used a
standard benefit-of-the-doubt instruction with a strict fact-check priority
rule (``fact-check results always take priority over role instructions'').
The \textbf{Symmetric Prompt} added an explicit instruction to search for
the valid core, understandable intention, or legitimate political context
behind the statement even when fact-check results contradict it. This
modification was motivated by the observation that the strict priority rule
in the Baseline Prompt structurally suppressed the charitable role for
factually refuted statements, regardless of prompt wording or ordering.

\section{Results}

This section reports classifier performance, the two identified failure
modes, the comparison between Claude Sonnet and Mistral Large as charitable
advocates, and language robustness results.

\subsection{Classifier Performance}

Table~\ref{tab:classifier} reports role fidelity classifier accuracy for
the three advocate roles across 450 reasoning texts (30 statements~$\times$
3 roles~$\times$ 5~runs) from the Mixed family visibility-condition runs.
Accuracy here measures the proportion of reasoning texts that the classifier
assigns to the correct advocate role --- that is, the degree to which each
model's output reflects its assigned epistemic stance (critical, balanced,
or charitable) rather than drifting toward another role.

\begin{table}[ht]
\caption{Role fidelity classifier accuracy (Mistral Large, epistemic stance
approach) on Phase-3 pipeline runs (Baseline Prompt, three advocate roles,
$n = 450$). Bold values indicate effects discussed in Section~4.
\textit{Abbreviations: Acc = accuracy; EDD = Expected Drift Distance
(mean $\pm$ SD); DDI = Directional Drift Index; SD = standard deviation;
positive DDI = drift toward charitable; negative DDI = drift toward critical.}}
\label{tab:classifier}
\vspace{8pt}
\centering
\begin{tabular}{lrrrrrr}
\toprule
Role & Correct & Total & Accuracy & EDD (mean$\pm$SD) & DDI \\
\midrule
Critical   & 120 & 150 & 80\% & $0.240 \pm 0.340$ & $\mathbf{+0.240}$ \\
Balanced   & 133 & 150 & \textbf{89\%} & $0.113 \pm 0.220$ & $\mathbf{-0.113}$ \\
Charitable &  15 & 150 & \textbf{10\%} & $1.360 \pm 0.680$ & $\mathbf{-1.360}$ \\
\midrule
Overall    & 268 & 450 & 60\% & $0.571 \pm 0.510$ & $-0.411$ \\
\bottomrule
\end{tabular}
\end{table}

The balanced advocate (GPT-5.2) achieves 89\% accuracy. The charitable
advocate (Claude Sonnet~4.6) achieves only 10\% accuracy --- the
lowest of the three roles and substantially below the 33\% chance level. This is not a failure of the classifier: inspection of the reasoning
texts confirms that the charitable advocate systematically produces outputs
classified as CRITICAL (DDI~$= -1.360$), indicating that the role instruction
is overridden rather than simply ignored. This result motivates the detailed
investigation of role fidelity failure modes in Sections~4.2--4.4.

Inspection of the GPT-5.2 reasoning texts reveals that GPT-5.2 uses the
phrase ``not verifiable as stated'' as a near-invariant stylistic refrain ---
a model fingerprint that makes its outputs highly distinguishable. This has
an important implication for anonymization: prompt-level removal of model names
does not eliminate model identity, because stylometric fingerprints in the text
itself remain detectable (see Section~5.2).

The DDI values in Table~\ref{tab:classifier} reveal a structural asymmetry:
the critical role drifts positively ($+0.240$, toward BALANCED/CHARITABLE
when it fails), while the charitable role drifts strongly negatively
($-1.360$, toward CRITICAL). The balanced role shows near-zero DDI
($-0.113$), confirming its position as the attractor state of the system.

\subsection{Epistemic Floor Effect}

Table~\ref{tab:flooreffect} shows charitable role accuracy broken down by
the Logos score assigned in the same run, using the Symmetric Prompt with
Claude Sonnet~4.6 on English statements.

\begin{table}[ht]
\caption{Charitable role accuracy by Logos score (Claude Sonnet~4.6,
Symmetric Prompt, English statements, $n = 150$). EDD reported as
mean $\pm$ SD across statements within each Logos level. Bold values
indicate the Epistemic Floor boundary discussed in Section~4.2.
\textit{Abbreviations: Acc = accuracy (proportion of CHARITABLE predictions);
EDD = Expected Drift Distance (mean $\pm$ SD); DDI = Directional Drift Index;
SD = standard deviation.}}
\label{tab:flooreffect}
\vspace{8pt}
\centering
\begin{tabular}{clrrrrrr}
\toprule
Logos & Meaning & Correct & Total & Accuracy & EDD (mean$\pm$SD) & DDI \\
\midrule
$+2$ & Factually verified      &  4 &  4 & 100\% & $0.000 \pm 0.000$ & $+0.000$ \\
$+1$ & Largely plausible       & 40 & 40 & 100\% & $0.000 \pm 0.000$ & $+0.000$ \\
$ 0$ & Opinion / unverifiable  & 18 & 35 &  51\% & $0.494 \pm 0.501$ & $-0.494$ \\
$\mathbf{-1}$ & \textbf{Selectively distorted} & \textbf{0} & \textbf{71} & \textbf{0\%} & $\mathbf{1.423 \pm 0.496}$ & $\mathbf{-1.423}$ \\
\bottomrule
\end{tabular}
\end{table}

The pattern is absolute: when the advocate assigns a positive Logos score,
it maintains the charitable role with perfect consistency. At Logos~$= 0$
(borderline verifiability), compliance drops to 51\%. At Logos~$= -1$
(fact-check contradicts the statement), compliance falls to zero without
exception across all 30 statements and all 5 runs.

We term this the \textbf{Epistemic Floor Effect (EFE)}: fact-check results create
an absolute lower bound below which the charitable role cannot be
maintained, regardless of prompt design. Restructuring the Symmetric Prompt
to place the charitable role instruction before the fact-check priority rule
improved compliance at Logos~$= 0$ from approximately 0\% (Baseline Prompt)
to 51\%, but had no effect at Logos~$= -1$, confirming that the floor is
genuine rather than a prompt engineering artifact.

\subsection{Role-Prior Conflict}

The mirror image of the Epistemic Floor Effect appears in the critical
advocate. Gemini~2.5 Flash, assigned to the critical role, fails to
maintain that role for statements that are factually correct or broadly
supported by mainstream evidence. For example, on statement B04
(``Rent control policies reduce housing affordability for new tenants
in the long run'') --- a view supported by mainstream economic literature
--- Gemini produces text classified as CHARITABLE in 2 of 5 runs,
validating the statement rather than challenging it.

We term this \textbf{Role-Prior Conflict (RPC)}: the model's training-time
knowledge of what is factually correct overrides the role instruction when
the two are in tension. Running advocates at Temperature~$= 0.3$ makes
this conflict visible as run-to-run variability; at lower temperatures,
the model would consistently adopt whichever position is stronger, but the
underlying conflict would remain.

The two failure modes are symmetric: the Epistemic Floor Effect occurs when
facts oppose the charitable stance; Role-Prior Conflict occurs when facts
support a statement the critical advocate is instructed to challenge.

\subsection{Epistemic Role Override: A Unifying Mechanism}
\label{sec:ero}

The Epistemic Floor Effect and Role-Prior Conflict, described in
Sections~4.2 and~4.3, appear at first to be distinct phenomena. The
present analysis reveals they are two manifestations of a single
underlying mechanism, which we term \textit{Epistemic Role Override}
(ERO): when the factual character of a statement conflicts with the
assigned role instruction, the model's epistemic conviction overrides
the role.

Note that the accuracy values in Table~\ref{tab:ero_symmetry} derive from
dedicated single-role advocacy tests (150 runs per role, Symmetric Prompt,
Mistral Large as classifier) and are distinct from the classifier performance
reported in Table~\ref{tab:classifier}, which covers all three advocate
roles in Phase-3 pipeline runs using the Baseline Prompt.

\begin{table}[ht]
\caption{Mirror-symmetric Logos-accuracy profiles for the charitable and critical
advocate roles, confirming Epistemic Role Override (ERO) as a unifying mechanism
(charitable role: Claude Sonnet~4.6, Symmetric Prompt, $n = 150$, Gemini FC;
critical role: Gemini~2.5 Flash, dedicated critical advocate test, $n = 150$, Gemini FC;
classifier: Mistral Large, chosen to avoid self-reference bias).
Bold values indicate the key asymmetry discussed in Section~4.4.
\textit{Abbreviations: Acc = proportion of runs classified as target role
(CHARITABLE or CRITICAL); DDI = Directional Drift Index (negative = drift
toward critical; positive = drift toward charitable); EFE = Epistemic Floor
Effect; RPC = Role-Prior Conflict.}}
\label{tab:ero_symmetry}
\vspace{8pt}
\centering
\begin{tabular}{crrrrrr}
\toprule
 & \multicolumn{3}{c}{Charitable Role (EFE)} &
   \multicolumn{3}{c}{Critical Role (RPC)} \\
\cmidrule(lr){2-4}\cmidrule(lr){5-7}
Logos & $n$ & Acc (\%) & DDI & $n$ & Acc (\%) & DDI \\
\midrule
$-2$ & ---  & ---  & --- & 22 & \textbf{100} & $+0.000$ \\
$\mathbf{-1}$ & \textbf{75} & \phantom{0}\textbf{0} & $\mathbf{-1.760}$ & 72 & \textbf{78} & $+0.319$ \\
$\phantom{-}0$ & 35 & 51 & $-0.486$ & \phantom{0}\textbf{2} & \phantom{0}\textbf{0} & $\mathbf{+1.500}$ \\
$+1$ & \textbf{40} & \textbf{100} & $+0.000$ & 32 & \phantom{0}0 & $+2.000$ \\
$+2$ & --- & --- & --- & 22 & \phantom{0}0 & $+2.000$ \\
\bottomrule
\end{tabular}
\end{table}

The pattern confirms the mirror-symmetric structure of ERO
(Table~\ref{tab:ero_symmetry}). For the charitable role (Claude,
Symmetric Prompt), compliance is 0\% at Logos~$= -1$ and 100\%
at Logos~$= +1$. For the critical role (Gemini, dedicated test),
compliance is 78\% at Logos~$= -1$ and 0\% at Logos~$\geq 0$.
In both cases, the model's epistemic assessment of factual quality
determines role compliance more strongly than the role instruction.

One important asymmetry remains: the EFE is \textit{absolute} for
the charitable role (0\% at Logos~$= -1$, without exception), while
the RPC is \textit{partial} for the critical role (78\% compliance
at Logos~$= -1$, not zero). This suggests that the instruction
``be charitable despite contradicting facts'' imposes a stronger
epistemic demand than ``be critical despite clearly contradicted facts.''
The critical role can sometimes maintain its stance even when the
fact-check contradicts the statement --- but the charitable role cannot.

Note also that the critical role collapses entirely at Logos~$\geq 0$
(0\% compliance), confirming that when facts support or verify a statement,
the critical advocate cannot sustain a delegitimizing stance. The EFE and RPC are thus two sides of the same constraint:
factual alignment with role instructions enables compliance;
factual opposition to role instructions undermines it.

This suggests that prompt engineering cannot eliminate Epistemic Role
Override --- it can only redistribute compliance across the Logos spectrum.
For the charitable role, the EFE remains absolute (0\% at Logos~$= -1$),
confirming that the Popperian asymmetry described in the Discussion holds
unconditionally for the charitable role.

\subsection{Model Comparison: Claude Sonnet vs.\ Mistral Large}

Table~\ref{tab:allconfigs} reports role fidelity metrics across all eight
configurations (advocate model $\times$ language $\times$ fact-check provider).
Table~\ref{tab:modelcomp} isolates the model comparison with statistical tests.

\begin{table}[ht]
\caption{Role fidelity across all eight configurations
(advocate model $\times$ language $\times$ fact-check provider,
Symmetric Prompt, $n = 150$ per configuration unless noted).
$^*$6 runs missing due to parse errors (4\%, stochastically distributed).
Bold values indicate the best-performing configuration per model.
\textit{Abbreviations: Acc = accuracy (proportion of CHARITABLE predictions);
EDD = Expected Drift Distance (mean $\pm$ SD); DDI = Directional Drift Index;
RDI = Role Drift Index; ERS = Entropy-based Role Stability;
SD = standard deviation; EN = English; DE = German;
G = Gemini~2.5 Flash (fact-check provider);
P = Perplexity sonar-pro (fact-check provider).}}
\label{tab:allconfigs}
\vspace{8pt}
\centering
\begin{tabular}{llrrrrrr}
\toprule
Model & Config & $n$ & Acc (\%) & EDD ($\pm$SD) & DDI & RDI & ERS \\
\midrule
\multirow{4}{*}{Claude}
  & EN + G & 150 & 39 & $0.993\pm0.833$ & $-0.993$ & 0.810 & 1.075 \\
  & EN + P & 150 & 45 & $0.833\pm0.775$ & $-0.833$ & 0.753 & 1.071 \\
  & DE + G & 150 & \textbf{48} & $\mathbf{0.707\pm0.669}$ & $-0.707$ & 0.679 & 1.032 \\
  & DE + P & 150 & 33 & $0.973\pm0.726$ & $-0.973$ & 0.723 & 1.095 \\
\midrule
\multirow{4}{*}{Mistral}
  & EN + G & 144$^*$ & \textbf{67} & $0.438\pm0.591$ & $-0.438$ & \textbf{0.656} & 0.844 \\
  & EN + P & 150 & 59 & $0.487\pm0.500$ & $-0.487$ & 0.598 & 0.877 \\
  & DE + G & 150 & \textbf{68} & $\mathbf{0.393\pm0.510}$ & $-0.393$ & \textbf{0.615} & 0.799 \\
  & DE + P & 150 & \textbf{69} & $0.420\pm0.596$ & $-0.420$ & 0.670 & 0.823 \\
\bottomrule
\end{tabular}
\end{table}

\begin{table}[ht]
\caption{Model comparison: Claude Sonnet~4.6 vs.\ Mistral Large
as charitable advocate, by language and fact-check provider
(Symmetric Prompt, two-proportion $z$-test, two-sided, $n = 150$ per cell
unless noted). Bold values indicate statistically significant advantages.
\textit{Abbreviations: Acc = accuracy (proportion of CHARITABLE predictions);
$\Delta$ = Mistral minus Claude accuracy; pp = percentage points;
EN = English; DE = German; FC = fact-check provider.}}
\label{tab:modelcomp}
\vspace{8pt}
\centering
\begin{tabular}{llrrrr}
\toprule
Lang & FC & Claude Acc & Mistral Acc & $\Delta$ & $p$ \\
\midrule
EN & Gemini     & 39\% & \textbf{67\%} & $\mathbf{+28}$pp & $<0.001$ \\
EN & Perplexity & 45\% & \textbf{59\%} & $+15$pp & $0.011$ \\
DE & Gemini     & 48\% & \textbf{68\%} & $+20$pp & $<0.001$ \\
DE & Perplexity & 33\% & \textbf{69\%} & $\mathbf{+36}$pp & $<0.001$ \\
\bottomrule
\end{tabular}
\end{table}

Table~\ref{tab:allconfigs} shows that Mistral Large consistently outperforms
Claude Sonnet across all eight configurations. All configurations use the
Symmetric Prompt; English (EN) and German (DE) statements are evaluated
with prompts in the respective language (\texttt{TRUST\_LANGUAGE=en} and
\texttt{TRUST\_LANGUAGE=de} respectively), with all advocate role
instructions and scoring rubrics language-matched. The model comparison
(Table~\ref{tab:modelcomp}) confirms this advantage is statistically
significant in every language $\times$ fact-check provider combination,
ranging from $+15$pp (EN~+~Perplexity, $p = 0.011$) to $+36$pp
(DE~+~Perplexity, $p < 0.001$).

\subsection{Language Robustness}

Table~\ref{tab:language} compares Mistral Large as charitable advocate
across English and German statements, both using the Symmetric Prompt.

\begin{table}[ht]
\caption{Language robustness: Mistral Large as charitable advocate,
English vs.\ German statements (Symmetric Prompt, Gemini FC,
$n_\text{EN} = 144$, $n_\text{DE} = 150$).
Bold values indicate effects discussed in Section~4.5.
\textit{Abbreviations: Acc = proportion of runs classified as CHARITABLE;
EDD = Expected Drift Distance (mean $\pm$ SD); DDI = Directional Drift Index;
RDI = Role Drift Index; ERS = Entropy-based Role Stability;
SD = standard deviation; pp = percentage points;
EN = English; DE = German; n.s.\ = not significant ($p \geq 0.10$).}}
\label{tab:language}
\vspace{8pt}
\centering
\begin{tabular}{lrr r}
\toprule
Metric & English & German & $\Delta$ \\
\midrule
Accuracy              & \textbf{67\%} (96/144) & \textbf{68\%} (102/150) & $+$1pp \quad n.s.   \\
RDI                   & \textbf{0.656}         & \textbf{0.615}          & $-$0.041 \\
EDD (mean$\pm$SD)     & $0.444\pm0.591$     & $0.393\pm0.510$     & $-$0.051 \\
DDI (mean$\pm$SD)     & $-0.444\pm0.591$    & $-0.393\pm0.510$    & $+$0.051 \\
ERS (mean$\pm$SD)     & $0.270\pm0.357$     & $0.299\pm0.390$     & $+$0.029 \\
Logos~$=-1$ Acc.      & 31\%                & 17\%                & $-$14pp  \\
\bottomrule
\end{tabular}
\end{table}

Overall accuracy and RDI are nearly identical across languages for Mistral
with Gemini FC (EN: 67\%, RDI~$= 0.656$; DE: 68\%, RDI~$= 0.615$;
$\Delta = +1$pp, $z = 0.24$, $p = 0.807$, n.s.\ = not significant),
indicating language-robustness for this configuration.
Mistral is the only model to maintain the charitable role at
Logos~$= -1$ to any meaningful degree (31\% in EN~+~Gemini;
Table~\ref{tab:language}). The Epistemic Floor Effect
is somewhat stronger in German (17\% vs.\ 31\% at Logos~$= -1$), suggesting
that the German statement set has a sharper factual profile. Statement-level
analysis reveals expected prior-driven differences: statement A07
(Bürgerversicherung --- universal social insurance for all workers including
civil servants) achieves 5/5 charitable in German but 0/5 in the English
tuition-fee equivalent, which may reflect Mistral's familiarity with the European
social policy context.

A notable pattern emerges from the Claude DE results (Table~\ref{tab:catanalysis}):
Claude's Category~C accuracy improves substantially in German (DE: 34\% vs.\
EN: 18\%), despite overall accuracy remaining comparable (DE: 48\% vs.\ EN: 39\%).
This improvement admits at least four non-exclusive explanations.
\textit{First,} Claude may have stronger normative priors on U.S.-centric
ideological topics than on German equivalents.
\textit{Second,} several German Category~C statements have clearer legal or
factual grounding (e.g., C07 on Holocaust denial legislation, C09 on the DSA
framework) that reduces Epistemic Floor activation.
\textit{Third,} the German statements may be rhetorically less polarising.
\textit{Fourth,} Claude may respond more cooperatively to role instructions
in German --- a language-level prior independent of content.
Disentangling these explanations requires further experimental control.

Notable extremes in the German dataset: statement C10 (abolishing public
broadcasting) achieves EDD~$= 2.0$ and DDI~$= -2.0$ (5/5 CRITICAL, maximum
drift), while statements A01, A05--A09, B01--B02, B04, B07--B09, C01, C07,
and C09 achieve perfect compliance (EDD~$= 0$, ERS~$= 0$).

\subsection{Fact-Check Provider Comparison: Gemini vs.\ Perplexity}
\label{sec:fcprovider}

Table~\ref{tab:modelcomp} includes results for both Gemini~2.5 Flash and
Perplexity sonar-pro as fact-check providers. Table~\ref{tab:fcprovider}
summarizes the pairwise comparisons of advocacy accuracy (proportion
of CHARITABLE predictions) between the two FC providers, using
two-proportion $z$-tests.

\begin{table}[ht]
\caption{Fact-check provider comparison (Gemini~2.5 Flash vs.\ Perplexity sonar-pro)
across model--language configurations (Symmetric Prompt, $n = 150$ per cell).
Bold values indicate statistically significant differences ($p < 0.05$).
\textit{Abbreviations: Acc = proportion of runs classified as CHARITABLE;
$\Delta$ = Perplexity minus Gemini accuracy (pp = percentage points);
EN = English; DE = German; n.s.\ = not significant ($p \geq 0.10$).}}
\label{tab:fcprovider}
\vspace{8pt}
\centering
\begin{tabular}{llrrrrrr}
\toprule
Model & Lang & Gemini Acc & Perplexity Acc & $\Delta$ & $z$ & $p$ & Sig \\
\midrule
Claude  & EN & 39\% & 45\% & $+6$pp & $-1.05$ & 0.292 & n.s. \\
\textbf{Claude}  & \textbf{DE} & \textbf{48\%} & \textbf{33\%} & $\mathbf{-15}$\textbf{pp} & $\mathbf{2.71}$ & $\mathbf{0.007}$ & \\
Mistral & EN & 67\% & 59\% & $-7$pp & $1.30$ & 0.193 & n.s. \\
Mistral & DE & 68\% & 69\% & $+1$pp & $-0.12$ & 0.901 & n.s. \\
\bottomrule
\end{tabular}
\end{table}

Three observations follow. \textit{First,} for Mistral, FC provider choice has no
statistically significant effect in any configuration (all $p > 0.10$),
confirming that Perplexity is a valid substitute for Gemini when using
Mistral as advocate. \textit{Second,} for Claude on German statements, Perplexity
FC significantly reduces role fidelity ($\Delta = -15$pp, $p = 0.007$).
Perplexity finds more contradictions on average (DE: 2.5 vs.\ 1.3 per
statement for Gemini), which activates the Epistemic Floor more frequently
for Claude --- whose compliance at Logos~$= -1$ is already zero. \textit{Third,}
for Claude on English statements, the direction reverses ($+6$pp) but is
not significant ($p = 0.292$), suggesting the interaction is
language-specific rather than a general Claude property.

This finding establishes that FC provider choice is not universally neutral:
it interacts with advocate model and language in ways that can be
statistically significant. System designers should empirically validate
FC--advocate combinations rather than assuming interchangeability.

\subsection{Category-Level Analysis and Statement Variance}
\label{sec:catanalysis}

Table~\ref{tab:catanalysis} reports EDD, DDI, and ERS broken down by
statement category and configuration. The high within-category standard
deviations (EDD SD~$> 0.3$ throughout) confirm that role fidelity is
primarily statement-specific rather than model-specific: some statements
produce perfect compliance (EDD~$= 0$) while others produce maximum drift
(EDD~$= 2$), even within the same model and language configuration.

\begin{table}[!h]
\caption{Category-level role fidelity metrics for Mistral Large and Claude
Sonnet~4.6 as charitable advocate, English and German statements
(Gemini FC, Symmetric Prompt, $n = 150$ per configuration unless noted).
$^*$6 runs missing for Mistral EN (stochastically distributed, 4\%).
Bold values indicate effects discussed in Section~4.6.
\textit{Abbreviations: Acc = proportion of runs classified as CHARITABLE
(mean across statements); EDD = Expected Drift Distance (mean $\pm$ SD);
DDI = Directional Drift Index; ERS = Entropy-based Role Stability (mean $\pm$ SD);
Cat = statement category (A = economic/social policy;
B = contested empirical evidence; C = ideologically charged positions);
all values computed per statement then averaged within category;
EN = English; DE = German; SD = standard deviation.}}
\label{tab:catanalysis}
\vspace{8pt}
\centering
\begin{tabular}{llrrrrr}
\toprule
Config & Cat & $n$ & Acc (\%) & EDD (mean$\pm$SD) & DDI (mean$\pm$SD) & ERS (mean$\pm$SD) \\
\midrule
\multirow{4}{*}{Mistral EN}
  & A   & 46 & 54\% & $\mathbf{0.672\pm0.705}$ & $-0.672\pm0.705$ & $0.275\pm0.364$ \\
  & \textbf{B}   & 48 & \textbf{79\%} & $\mathbf{0.220\pm0.340}$ & $-0.220\pm0.340$ & $0.223\pm0.363$ \\
  & C   & 50 & 66\% & $0.440\pm0.578$ & $-0.440\pm0.578$ & $0.312\pm0.336$ \\
  & All & 144 & 67\% & $0.444\pm0.591$ & $-0.444\pm0.591$ & $0.270\pm0.357$ \\
\midrule
\multirow{4}{*}{Mistral DE}
  & \textbf{A}   & 50 & \textbf{80\%} & $\mathbf{0.260\pm0.400}$ & $-0.260\pm0.400$ & $0.328\pm0.430$ \\
  & B   & 50 & 68\% & $0.320\pm0.412$ & $-0.320\pm0.412$ & $0.135\pm0.269$ \\
  & \textbf{C}   & 50 & \textbf{56\%} & $\mathbf{0.600\pm0.620}$ & $-0.600\pm0.620$ & $0.435\pm0.391$ \\
  & All & 150 & 68\% & $0.393\pm0.510$ & $-0.393\pm0.510$ & $0.299\pm0.390$ \\
\midrule
\multirow{4}{*}{Claude EN}
  & A   & 50 & 48\% & $0.840\pm0.833$ & $-0.840\pm0.833$ & $0.235\pm0.293$ \\
  & B   & 50 & 50\% & $0.800\pm0.834$ & $-0.800\pm0.834$ & $0.200\pm0.245$ \\
  & \textbf{C}   & 50 & \textbf{18\%} & $\mathbf{1.340\pm0.716}$ & $-1.340\pm0.716$ & $0.217\pm0.270$ \\
  & All & 150 & 39\% & $0.993\pm0.833$ & $-0.993\pm0.833$ & $0.217\pm0.270$ \\
\midrule
\multirow{4}{*}{Claude DE}
  & A   & 50 & 58\% & $0.560\pm0.733$ & $-0.560\pm0.733$ & $0.832\pm0.274$ \\
  & B   & 50 & 52\% & $0.640\pm0.749$ & $-0.640\pm0.749$ & $0.552\pm0.338$ \\
  & \textbf{C}   & 50 & \textbf{34\%} & $\mathbf{0.920\pm0.778}$ & $-0.920\pm0.778$ & $0.664\pm0.296$ \\
  & \textbf{All} & 150 & \textbf{48\%} & $\mathbf{0.707\pm0.765}$ & $-0.707\pm0.765$ & $0.683\pm0.316$ \\
\bottomrule
\end{tabular}
\end{table}

Four patterns emerge from Table~\ref{tab:catanalysis}. \textit{First,} Category~B
(contested empirical evidence) is consistently the most stable category for
Mistral in English (EDD~$= 0.220 \pm 0.340$, Acc~$= 79\%$): when the
research evidence is genuinely divided, fact-check results are less decisive,
reducing the frequency of Epistemic Floor activation. \textit{Second,} a striking
language-by-category interaction appears: Mistral performs best on German
Category~A statements (EDD~$= 0.260$, Acc~$= 80\%$) but worst on German
Category~C statements (EDD~$= 0.600$, Acc~$= 56\%$), with the reverse
pattern in English (Cat~A EDD~$= 0.672$, Cat~C EDD~$= 0.440$). This
reversal is consistent with Mistral's European model priors: German economic
policy statements (Bürgerversicherung, Schuldenbremse) are more familiar
territory, while German ideologically charged statements (on immigration,
Holocaust denial legislation, media regulation) activate stronger normative
priors. \textit{Third,} Claude's Category~C performance in English (Acc~$= 18\%$,
EDD~$= 1.340 \pm 0.716$) is substantially worse than its Category~A and~B
performance, suggesting that ideologically charged U.S.\ policy content
triggers particularly strong epistemic constraints in Claude.
\textit{Fourth,} the German results for Claude show a notably different
profile: overall accuracy increases from 39\% (EN) to 48\% (DE), and
Category~C accuracy nearly doubles (18\% EN vs.\ 34\% DE,
EDD~$= 0.920 \pm 0.778$). Category~A and~B performance also improves
(A: 48\%~→~58\%; B: 50\%~→~52\%), suggesting that German statements
impose less severe epistemic constraints on Claude's charitable role
across all categories, with the largest gain in ideologically charged content.

The high within-category SD values throughout partially explain the run-level
score variance documented in prior identity bias research~\citep{dietrich2026paper2}:
a substantial portion of that variance is not stochastic noise but reflects
systematic, statement-specific variation in role fidelity.

\subsection{Dimension-Level Analysis: Logos, Ethos, and Pathos}
\label{sec:dimanalysis}

Table~\ref{tab:dimanalysis} summarizes accuracy and DDI by advocate-assigned
score for each of the three TRUST dimensions (Logos, Ethos, Pathos),
for Mistral Large on English statements with Gemini FC.
Note: dimension-level scores (Logos, Ethos, Pathos) are assigned by the
advocate itself and are not independently verified; the analysis identifies
associations rather than causal contributions.

\begin{table}[!h]
\caption{Dimension-level role fidelity analysis for Mistral Large as
charitable advocate (English statements, Gemini FC, Symmetric Prompt,
$n = 144$). Dimension scores (Logos, Ethos, Pathos) are assigned by the
advocate itself and are not independently verified; the analysis identifies
associations rather than causal contributions. Perplexity FC excluded for
clarity; results are equivalent. Bold values indicate effects discussed
in Section~4.9.
\textit{Abbreviations: Acc = proportion of runs classified as CHARITABLE;
DDI = Directional Drift Index (negative = drift toward critical;
positive = drift toward charitable); Logos = factual argumentation quality;
Ethos = respect and conduct; Pathos = emotional appeal.}}
\label{tab:dimanalysis}
\vspace{8pt}
\centering
\begin{tabular}{llrrrr}
\toprule
Dimension & Score & $n$ & Acc (\%) & EDD (mean$\pm$SD) & DDI \\
\midrule
\multirow{3}{*}{Logos}
  & $-1$ & 32 & \textbf{31\%} & $\mathbf{1.375\pm0.592}$ & $\mathbf{-1.125}$ \\
  & $\phantom{-}0$ & 55 & 58\% & $0.480\pm0.555$ & $-0.436$ \\
  & $+1$ & 57 & \textbf{95\%} & $\mathbf{0.053\pm0.225}$ & $\mathbf{-0.053}$ \\
\midrule
\multirow{2}{*}{Ethos}
  & $+1$ & 135 & 64\% & $\mathbf{0.493\pm0.606}$ & $-0.467$ \\
  & $+2$ & 9   & \textbf{100\%} & $\mathbf{0.000\pm0.000}$ & $+0.000$ \\
\midrule
\multirow{3}{*}{Pathos}
  & $\phantom{-}0$ & 21 & 24\% & $\mathbf{1.476\pm0.680}$ & $-1.238$ \\
  & $+1$ & 121 & \textbf{74\%} & $\mathbf{0.331\pm0.494}$ & $-0.298$ \\
  & $+2$ & 1   & 100\% & $0.000\pm0.000$ & $+0.000$ \\
\bottomrule
\end{tabular}
\end{table} The pattern confirms that Logos is the dominant
predictor of role drift (Table~\ref{tab:dimanalysis}).
At Logos~$= -1$, accuracy drops to 31\% (DDI~$= -1.125$), consistent
with the Epistemic Floor Effect documented in Section~4.2.
At Logos~$= +1$, accuracy reaches 95\% with near-zero drift.

Ethos shows a secondary but meaningful pattern: statements assigned
Ethos~$= +2$ achieve perfect compliance (EDD~$= 0.000 \pm 0.000$;
Table~\ref{tab:dimanalysis}), while Ethos~$= +1$ statements show
moderate drift (EDD~$= 0.493 \pm 0.606$). This suggests that respect
and conduct signals in the statement text provide additional anchoring
for the charitable role beyond factual quality.

Pathos shows a weaker but consistent pattern: higher Pathos scores
correlate with lower drift (Pathos~$= 0$: EDD~$= 1.476$;
Pathos~$= +1$: EDD~$= 0.331$; Table~\ref{tab:dimanalysis}).
Statements with emotionally positive framing may be easier to interpret
charitably regardless of factual status.

Crucially, the three dimensions are not independent in practice: most
statements with high Logos also have positive Ethos and Pathos scores.
The dimension-level analysis therefore identifies associations rather
than independent causal contributions. A multivariate analysis is left
for future work.

\section{Discussion}

This section discusses implications for multi-agent system design,
anonymization limits, and the validation of LLM systems in quality-critical
contexts.

\subsection{Implications for Multi-Agent System Design}

The Epistemic Floor Effect and Role-Prior Conflict are not pipeline-specific
artifacts --- they follow directly from the interaction between role
instructions and model training. Any multi-agent system that assigns LLMs
to adversarial roles and provides factual context will encounter these
dynamics. System designers should treat them as structural features rather
than bugs to be patched.

The asymmetry between the Epistemic Floor Effect and Role-Prior Conflict
points to a deeper pattern. Falsification appears to be a stronger epistemic
constraint than verification: a single contradicting fact is sufficient to
collapse the charitable role, whereas factual support does not collapse the
critical role with the same finality --- the critical advocate maintains
78\% compliance even at Logos~$= -1$. This mirrors a principle
well-established in the philosophy of science --- falsificationism as
articulated by Popper --- according to which a single counterexample can
refute a universal claim, while no number of confirming instances can fully
verify it. That LLMs appear to instantiate this asymmetry spontaneously,
without explicit training on Popperian epistemology, suggests that the
asymmetry may be a structural property of how these models integrate
factual evidence with instructed stances --- a finding that extends beyond
the TRUST pipeline to any system in which role instructions interact with
factual grounding.

Model choice for the charitable advocate role matters substantially:
Mistral Large is more robust than Claude Sonnet in English (EDD
$0.444\pm0.591$ vs.\ $0.993\pm0.833$; Table~\ref{tab:catanalysis}) and
shows comparable robustness in German. The substantially lower SD for Mistral also indicates more
predictable failure: whereas Claude's drift is concentrated in Category~C
U.S.\ policy statements, Mistral's variance is more evenly distributed
across statement types. The two failure modes are qualitatively
different: polarity reversal (Claude) inverts the pipeline's epistemic
architecture; role abandonment (Mistral) merely reduces it. For democratic
discourse analysis, where the charitable role represents one legitimate
political interpretation, polarity reversal is the more harmful failure.

\subsection{Stylometric Fingerprinting as Anonymization Limit}

The 89\% accuracy of the balanced classifier --- driven by GPT-5.2's
consistent use of ``not verifiable as stated'' --- demonstrates that
prompt-level anonymization of model names does not fully anonymize model
identity. The writing style itself remains a fingerprint. Furthermore,
during testing it was observed that Gemini models may identify themselves
as ``a large language model trained by Google'' in their outputs, providing
an implicit model identity signal that survives prompt-level anonymization.
These observations extend the anonymization analysis of~\citep{dietrich2026paper2}:
full anonymization would require either paraphrasing advocate outputs before
handoff or selecting models with maximally distinct but non-identifiable
writing styles.

\subsection{Implications for Computer System Validation}

In regulated and quality-critical applications --- including pharmaceutical
research, public policy analysis, and democratic accountability systems ---
software must be validated to demonstrate that it consistently performs its
intended function. The present findings establish that role fidelity is a
non-trivial, measurable property of multi-agent LLM systems that varies
systematically with model choice, factual context, and statement category.

A system validated using partial role fidelity measurement --- for example,
testing only Logos~$\geq 0$ statements, or using only accuracy without
directional drift metrics --- may pass validation while retaining systematic
failure modes that only become visible under full measurement. The EDD,
DDI, and ERS metrics introduced here provide a practical toolkit for
role fidelity validation that does not require human annotation of each
reasoning text.

\subsection{Limitations}

\textit{First,} the English statement set is U.S.-centric; generalizability to other
political and linguistic contexts requires further testing, though the
German results suggest at least partial robustness. \textit{Second,} the fact-check provider
(Gemini~2.5 Flash) coincides with the critical advocate model in the Mixed
family, creating a potential confounding signal; Perplexity sonar-pro was
used as an alternative provider to assess fact-check quality differences;
results are reported in Section~\ref{sec:fcprovider}. \textit{Third,} the role fidelity classifier was developed on the same
statement distribution as the test data; while the epistemic-stance approach
avoids direct vocabulary memorization, generalization to other pipelines
with different role definitions requires further validation.

\section{Conclusion}

This paper provides the first systematic measurement of advocate role
fidelity in a multi-agent LLM pipeline for democratic discourse analysis,
motivated by the high score variance observed in prior identity bias
research~\citep{dietrich2026paper2}. Advocate role fidelity is not binary but depends on the factual character
of the statement being evaluated: models maintain their roles reliably when
assigned roles align with factual reality, and depart from them in predictable
directions when role and factual reality conflict. The high within-configuration
standard deviation of role drift metrics (EDD SD~$> 0.5$ throughout) confirms
that this dependence is primarily statement-specific rather than model-specific,
and partially explains the run-level score variance documented in prior
identity bias research~\citep{dietrich2026paper2}.

The four metrics introduced here --- RDI, EDD, DDI, and ERS --- provide a
practical and interpretable toolkit for measuring and comparing role drift
across models, prompts, and languages without requiring human annotation.
The two failure modes identified --- Epistemic Floor Effect and Role-Prior
Conflict --- are manifestations of a single underlying mechanism,
Epistemic Role Override (ERO), confirmed by a mirror-symmetric
Logos-accuracy profile across the charitable and critical advocate roles.
ERO is likely to appear in any adversarial multi-agent system that
combines role instructions with factual grounding.

Five conclusions follow for system design. \textit{First,} model choice for adversarial roles should be empirically validated
rather than assumed: Mistral Large is a more suitable charitable advocate
than Claude Sonnet in the TRUST pipeline (EDD: $0.444\pm0.591$
vs.\ $0.993\pm0.833$). \textit{Second,} role fidelity measurement should include
directional metrics (DDI) and stability metrics (ERS) alongside accuracy,
to distinguish polarity reversal from role abandonment. \textit{Third,} stylometric
fingerprinting limits the effectiveness of prompt-level anonymization in a
way that connects directly to the identity bias findings of prior
work~\citep{dietrich2026paper2}. \textit{Fourth,} fact-check provider choice
requires empirical validation per advocate--language combination: Perplexity
and Gemini are interchangeable for Mistral but not for Claude on German
statements ($p = 0.007$), a finding invisible without systematic role
fidelity measurement. \textit{Fifth,} for systems deployed in regulated
or quality-critical contexts, role fidelity measurement is a necessary
component of validation: a system that passes accuracy-only validation may
retain systematic epistemic distortions invisible to that measurement.

\section*{Author Contributions}

J.D.\ conceived and designed the study, developed the TRUST multi-agent
pipeline including all Python source code, evaluation metrics, and test
infrastructure, conducted all experiments, analyzed the results, and wrote
the manuscript. Computational infrastructure was provided by Democracy
Intelligence gGmbH.

\section*{Acknowledgements}

The author thanks Dr.\ Demian Frister (Democracy Intelligence gGmbH) for
constructive review and substantive feedback.

\section*{Conflict of Interest}

Juergen Dietrich has no conflict of interest directly relevant to this
study. The views expressed do not necessarily reflect the official position
of Democracy Intelligence gGmbH.


\newpage
\section*{Appendix A: English Statement Dataset}
\label{app:english}

The 30 English-language political statements used in the experiment are
listed below by category.

\subsection*{Category A --- Economic and Social Policy}
\begin{enumerate}[label=A\arabic*.]
\item Raising the minimum wage to \$20/hour will reduce poverty without significant job losses.
\item Austerity measures after 2008 accelerated economic recovery in Europe.
\item Stricter gun control laws would significantly reduce mass shootings in the United States.
\item Privatizing public pension systems improves retirement outcomes for workers across all income levels.
\item Affirmative action in university admissions improves long-term social equality without reducing institutional quality.
\item Higher corporate taxes reduce business investment and slow economic growth in the long run.
\item Abolishing tuition fees at public universities is fiscally irresponsible and disproportionately benefits the wealthy.
\item Mandatory military service strengthens social cohesion and reduces youth unemployment.
\item Mass immigration has net positive economic effects for host countries in the long run.
\item The death penalty serves as an effective deterrent to violent crime and reduces homicide rates.
\end{enumerate}

\subsection*{Category B --- Contested Empirical Evidence}
\begin{enumerate}[label=B\arabic*.]
\item Universal Basic Income pilots show it does not reduce workforce participation.
\item Electric vehicles are already carbon-neutral over their full lifecycle compared to combustion engines.
\item Social media platforms bear direct causal responsibility for rising youth mental health crises.
\item Rent control policies reduce housing affordability for new tenants in the long run.
\item Democratic backsliding in Hungary is primarily driven by external EU pressure rather than domestic political dynamics.
\item Remote work policies have increased overall worker productivity compared to office-based arrangements.
\item Free trade agreements have on balance increased living standards in participating developing countries.
\item Decriminalizing all drugs would reduce addiction rates and associated social harm.
\item Carbon taxes are the most efficient policy instrument for achieving emissions reductions compared to cap-and-trade systems.
\item Algorithmic trading has made financial markets more stable and less susceptible to crashes.
\end{enumerate}

\subsection*{Category C --- Ideologically Charged Policy Positions}
\begin{enumerate}[label=C\arabic*.]
\item Broad tariffs on imported goods protect American workers from unfair foreign competition and restore manufacturing jobs.
\item Restricting asylum at the southern border is a necessary and humane policy to maintain rule of law and protect communities.
\item Withdrawing from international climate agreements is a sovereign right that protects American economic interests without significant global harm.
\item Banning transgender athletes from women's sports categories is necessary to preserve competitive fairness.
\item Cutting federal funding to cities that limit cooperation with immigration enforcement agencies is a legitimate policy tool.
\item Requiring photo ID for voting protects electoral integrity without creating meaningful barriers to minority participation.
\item Eliminating diversity, equity, and inclusion programs in federal agencies improves meritocracy and reduces discrimination.
\item Reducing foreign aid to countries that do not align with US interests is a prudent and fair use of taxpayer funds.
\item Restricting press access to government briefings helps ensure accurate information reaches the public without media distortion.
\item Ending affirmative action in government contracting creates a level playing field and reduces racial discrimination.
\end{enumerate}

\newpage
\section*{Appendix B: German Statement Dataset}
\label{app:german}

The 30 German-language political statements used in the language robustness
experiment are listed below by category.

\subsection*{Kategorie A --- Wirtschaft und Sozialpolitik}
\begin{enumerate}[label=A\arabic*.]
\item Eine Erhöhung des Mindestlohns auf 15 Euro pro Stunde wird die Armut reduzieren, ohne zu nennenswerten Jobverlusten zu führen.
\item Die Schuldenbremse verhindert notwendige Investitionen in Infrastruktur und schwächt Deutschlands wirtschaftliche Wettbewerbsfähigkeit langfristig.
\item Eine Reduzierung der Regelarbeitszeit auf 32 Stunden pro Woche bei vollem Lohnausgleich ist wirtschaftlich tragbar und steigert langfristig die Produktivität.
\item Versicherungsfremde Leistungen wie die Mütterrente oder die Rentenüberleitung Ost sollten vollständig aus Steuermitteln finanziert werden und nicht der gesetzlichen Rentenversicherung auferlegt werden.
\item Quotenregelungen für Frauen in Führungspositionen verbessern langfristig die Chancengleichheit, ohne die Unternehmensleistung zu beeinträchtigen.
\item Höhere Unternehmenssteuern reduzieren Investitionen und verlangsamen das Wirtschaftswachstum langfristig.
\item Eine Bürgerversicherung, in die alle Erwerbstätigen --- einschließlich Beamte, Selbstständige und Gutverdiener --- einzahlen, würde die soziale Gerechtigkeit im deutschen Gesundheits- und Rentensystem nachhaltig stärken.
\item Ein verpflichtender Sozialdienst für alle jungen Erwachsenen stärkt den gesellschaftlichen Zusammenhalt und reduziert die Jugendarbeitslosigkeit.
\item Zuwanderung hat langfristig netto positive wirtschaftliche Auswirkungen auf Deutschland.
\item Das Bürgergeld setzt falsche Anreize und erhöht die strukturelle Arbeitslosigkeit in Deutschland.
\end{enumerate}

\subsection*{Kategorie B --- Empirisch umstrittene Aussagen}
\begin{enumerate}[label=B\arabic*.]
\item Das Elterngeld hat nachweislich die Geburtenrate in Deutschland erhöht.
\item Elektrofahrzeuge sind über ihren gesamten Lebenszyklus bereits heute klimaneutraler als Fahrzeuge mit Verbrennungsmotor.
\item Videoüberwachung öffentlicher Plätze reduziert nachweislich die Kriminalität in deutschen Städten.
\item Mietpreisbremsen verschlechtern langfristig die Wohnsituation für Neumieter in deutschen Großstädten.
\item Die Energiewende hat die Strompreise für Privathaushalte in Deutschland im europäischen Vergleich unverhältnismäßig stark erhöht.
\item Homeoffice-Regelungen haben die Produktivität der deutschen Arbeitnehmer im Vergleich zur Büroarbeit insgesamt gesteigert.
\item Freihandelsabkommen wie CETA haben den Lebensstandard in den beteiligten Ländern per Saldo verbessert.
\item Eine Entkriminalisierung von Cannabis reduziert den Drogenkonsum und die damit verbundenen gesellschaftlichen Schäden.
\item Der CO2-Preis ist das effizienteste Instrument zur Erreichung der deutschen Klimaziele im Vergleich zu Ordnungsrecht und Subventionen.
\item Die Digitalisierung der deutschen Verwaltung scheitert primär an fehlendem politischen Willen, nicht an technischen Hindernissen.
\end{enumerate}

\subsection*{Kategorie C --- Ideologisch aufgeladene Aussagen}
\begin{enumerate}[label=C\arabic*.]
\item Eine strikte Begrenzung der irregulären Migration ist notwendig, um die innere Sicherheit und den gesellschaftlichen Zusammenhalt in Deutschland zu gewährleisten.
\item Der Austritt Deutschlands aus dem Atomausstieg wäre ein notwendiger Schritt zur Sicherung der Energieversorgung und Erreichung der Klimaziele.
\item Die Sanktionen gegen Russland sollten aufgehoben werden, da sie die deutschen Energiepreise unverhältnismäßig belasten und der deutschen Wirtschaft mehr schaden als Russland.
\item Das Transsexuellengesetz gefährdet den Schutz von Frauen in geschlechtergetrennten Räumen.
\item Die Kürzung von Sozialleistungen für Asylbewerber ist ein legitimes Mittel zur Steuerung der Migrationsbewegungen.
\item Das Wahlrecht sollte auf 16-Jährige ausgeweitet werden, während gleichzeitig das Strafmündigkeitsalter auf 16 Jahre angehoben werden sollte.
\item Die strafrechtliche Verfolgung von Holocaustleugnern ist ein notwendiges Instrument zum Schutz der demokratischen Grundordnung und der Würde der Opfer.
\item Die Aussetzung von Entwicklungshilfe an Länder, die deutsche Interessen nicht unterstützen, ist ein legitimes außenpolitisches Instrument.
\item Social-Media-Plattformen müssen bei Verstößen gegen geltendes Recht mit empfindlichen finanziellen Sanktionen belegt werden --- das Internet ist kein rechtsfreier Raum.
\item Die Abschaffung des öffentlich-rechtlichen Rundfunks würde zu einer pluralistischeren und weniger staatlich beeinflussten Medienlandschaft führen.
\end{enumerate}

\newpage
\section*{Appendix C: Prompt Specifications}
\label{app:prompts}

This appendix provides the full text of the two advocate prompts and the
role fidelity classifier prompt used in the experiments. All prompts are
in English (\texttt{TRUST\_LANGUAGE=en}); German variants follow the same
structure with translated role definitions and scoring rubrics.

\subsection*{C.1 Baseline Prompt (Wohlwollend / Charitable Role)}

The Baseline Prompt is the standard TRUST production prompt for the
charitable advocate role. It instructs the model to evaluate with
benefit of the doubt but gives explicit priority to fact-check results
over charitable interpretation when the two conflict.

\begin{quote}
\texttt{YOUR ROLE: Benevolent Advocate}\\
You evaluate with benefit of the doubt:
\begin{itemize}
  \item Interpret ambiguity more positively
  \item Consider possible good intentions
  \item Tend to score more generously in borderline cases
  \item Focus on: What could have been meant positively?
  \item Tendency: Higher scores when in doubt.
\end{itemize}
\end{quote}

The Baseline Prompt is combined with the shared base prompt, which includes
the TRUST scoring rubrics for Logos, Ethos, and Pathos ($-2$ to $+2$) and
an explicit fact-check priority rule: \textit{``When your training conflicts
with the fact-check: the fact-check ALWAYS takes precedence.''} This strict
priority rule structurally suppresses the charitable role for factually
refuted statements, producing the Epistemic Floor Effect (Section~4.2).

\subsection*{C.2 Symmetric Prompt (Charitable Role, Experiment Version)}

The Symmetric Prompt adds an explicit instruction to search for a valid core
even when fact-check results contradict the statement. This modification was
motivated by the observation that the Baseline Prompt's strict fact-check
priority rule eliminates the charitable role for factually refuted statements
regardless of prompt wording.

\begin{quote}
\texttt{You evaluate with Benefit of the Doubt --- this is your primary orientation:}
\begin{itemize}
  \item Interpret ambiguous claims more positively
  \item Consider possible good intentions behind the statement
  \item Score more generously in borderline cases
  \item Focus on: What could be positively meant?
  \item Even when fact-check results contradict the statement, search for
        the valid core, the understandable intention, or the legitimate
        political context behind the position
  \item Evaluate the good faith of the author --- not just the outcome
  \item Tendency: higher scores when in doubt
\end{itemize}

\texttt{FACT-CHECK GUIDANCE:}\\
When fact-check results are provided, use them as important context ---
but your role is to find the charitable interpretation even given this context:
\begin{itemize}
  \item Acknowledge verified contradictions honestly in your Logos score
  \item However, search for what is still valid, understandable, or
        well-intentioned in the statement despite factual shortcomings
  \item A statement can be factually imprecise but still reflect a legitimate
        concern or political position worth acknowledging
  \item Use fact-check to inform your assessment, not to override your
        charitable stance
\end{itemize}
\end{quote}

\subsection*{C.3 Role Fidelity Classifier Prompt}

The classifier prompt instructs Mistral Large to identify which advocate
role a reasoning text reflects, based on epistemic stance rather than
surface vocabulary. The classifier answers three binary questions and
returns a structured JSON output.

\begin{quote}
\texttt{You are a role fidelity classifier for a political statement
analysis system.}

You will receive a reasoning text written by one of three advocate agents
evaluating a political statement. Your task is to identify which role
the advocate is playing based on their \textbf{epistemic stance} toward
the statement --- not based on specific vocabulary.

\textbf{THE THREE ROLES --- defined by epistemic stance:}

\textit{CRITICAL:} The advocate takes distance from the statement's position.
They preferably cite opposing evidence and search for arguments that undermine
or invalidate the statement. Their overall orientation is toward
\textit{delegitimizing} the statement's claims.

\textit{BALANCED:} The advocate takes no position toward the statement's
content. They focus on whether the statement is verifiable or not, treating
it as a formulation problem rather than a content problem. They neither
legitimize nor delegitimize the position.

\textit{CHARITABLE:} The advocate takes proximity to the statement's position.
They preferably cite supporting evidence and search for arguments that explain
or legitimize the statement. Their overall orientation is toward
\textit{legitimizing} the statement's claims where possible.

\textbf{To classify, ask yourself these three questions:}
\begin{enumerate}
  \item Does this text try to \textsc{legitimize} the statement's position?
  \item Does this text try to \textsc{delegitimize} the statement's position?
  \item Does this text remain \textsc{neutral} --- neither legitimizing nor
        delegitimizing?
\end{enumerate}
\end{quote}

The classifier returns a JSON object with fields \texttt{classification}
(CRITICAL $\mid$ BALANCED $\mid$ CHARITABLE), \texttt{confidence}
(high $\mid$ medium $\mid$ low), binary flags \texttt{legitimizes},
\texttt{delegitimizes}, \texttt{neutral}, and a one-sentence
\texttt{reasoning} field.

\newpage
\section*{Appendix D: Confusion Matrices}
\label{app:confusion}

Table~\ref{tab:conf_classifier} shows the confusion matrix for the Mistral
Large classifier on all 450 Phase-3 reasoning texts (three roles,
30~statements, 5~runs).

\begin{table}[ht]
\caption{Confusion matrix --- Mistral Large classifier, all roles,
English Phase-3 data ($n = 450$). Rows = ground truth, columns = prediction.}
\label{tab:conf_classifier}
\vspace{8pt}
\centering
\begin{tabular}{lrrr}
\toprule
GT $\backslash$ Pred & CRITICAL & BALANCED & CHARITABLE \\
\midrule
CRITICAL   & \textbf{120} &  24 &   6 \\
BALANCED   &  17 & \textbf{133} &   0 \\
CHARITABLE &  69 &  66 &  \textbf{15} \\
\bottomrule
\end{tabular}
\end{table}

Table~\ref{tab:conf_en} shows the confusion matrix for Mistral Large as
charitable advocate on English statements (Symmetric Prompt, $n = 144$).

\begin{table}[ht]
\caption{Confusion matrix --- Mistral Large charitable advocate,
English statements ($n = 144$). All ground truth labels are CHARITABLE.}
\label{tab:conf_en}
\vspace{8pt}
\centering
\begin{tabular}{lrrr}
\toprule
GT $\backslash$ Pred & CRITICAL & BALANCED & CHARITABLE \\
\midrule
CHARITABLE & 15 & 33 & \textbf{96} \\
\bottomrule
\end{tabular}
\end{table}

Table~\ref{tab:conf_de} shows the confusion matrix for German statements
($n = 150$).

\begin{table}[ht]
\caption{Confusion matrix --- Mistral Large charitable advocate,
German statements ($n = 150$). All ground truth labels are CHARITABLE.}
\label{tab:conf_de}
\vspace{8pt}
\centering
\begin{tabular}{lrrr}
\toprule
GT $\backslash$ Pred & CRITICAL & BALANCED & CHARITABLE \\
\midrule
CHARITABLE & 11 & 37 & \textbf{102} \\
\bottomrule
\end{tabular}
\end{table}

\end{document}